\pdfoutput=1

\documentclass[11pt]{article}

\usepackage{enumitem}
\usepackage[utf8]{inputenc}
\usepackage[T1]{fontenc}

\usepackage{mdframed}
\usepackage{hyperref}
\usepackage{booktabs}

\newlist{inlinelist}{enumerate*}{1}
\setlist*[inlinelist,1]{%
  label=(\alph*),
}

\newlist{questions}{enumerate*}{1}
\setlist*[questions,1]{label=RQ\arabic*.,ref=RQ\arabic*}

\usepackage[final]{acl}

\usepackage{times}
\usepackage{latexsym}

\usepackage[T1]{fontenc}

\usepackage[utf8]{inputenc}

\usepackage{microtype}

\usepackage{inconsolata}

\usepackage{amsmath}
\usepackage{xcolor}

\usepackage{hyperref}
\usepackage{algorithm}
\usepackage{algpseudocode}
\usepackage{graphicx}
\usepackage{subcaption} 
\usepackage[normalem]{ulem}

\newcommand{\kt}{Kendall's~$\tau_c$~}

\title{Extrinsic Evaluation of Cultural Competence in Large Language Models}

\author{Shaily Bhatt \\
  Carnegie Mellon University  \\
  \texttt{shaily@cmu.edu} \\\And
  Fernando Diaz \\
  Carnegie Mellon University  \\
  \texttt{diazf@acm.org} \\}

\begin{document}
\maketitle

\begin{abstract}
Productive interactions between diverse users and language technologies require outputs from the latter to be culturally relevant and sensitive. Prior works have evaluated models' knowledge of cultural norms, values, and artefacts, without considering how this knowledge manifests in downstream applications. In this work, we focus on extrinsic evaluation of cultural competence in two text generation tasks, open-ended question answering and story generation. We quantitatively and qualitatively evaluate model outputs when an explicit cue of culture, specifically nationality, is perturbed in the prompts. Although we find that model outputs do vary when varying nationalities and feature culturally relevant words, we also find weak correlations between text similarity of outputs for different countries and the cultural values of these countries. Finally, we discuss important considerations in designing comprehensive evaluation of cultural competence in user-facing tasks.
\end{abstract}
\section{Introduction}
\label{sec:1-introduction}

\begin{figure*}[ht]
    \centering
    \includegraphics[width=\linewidth]{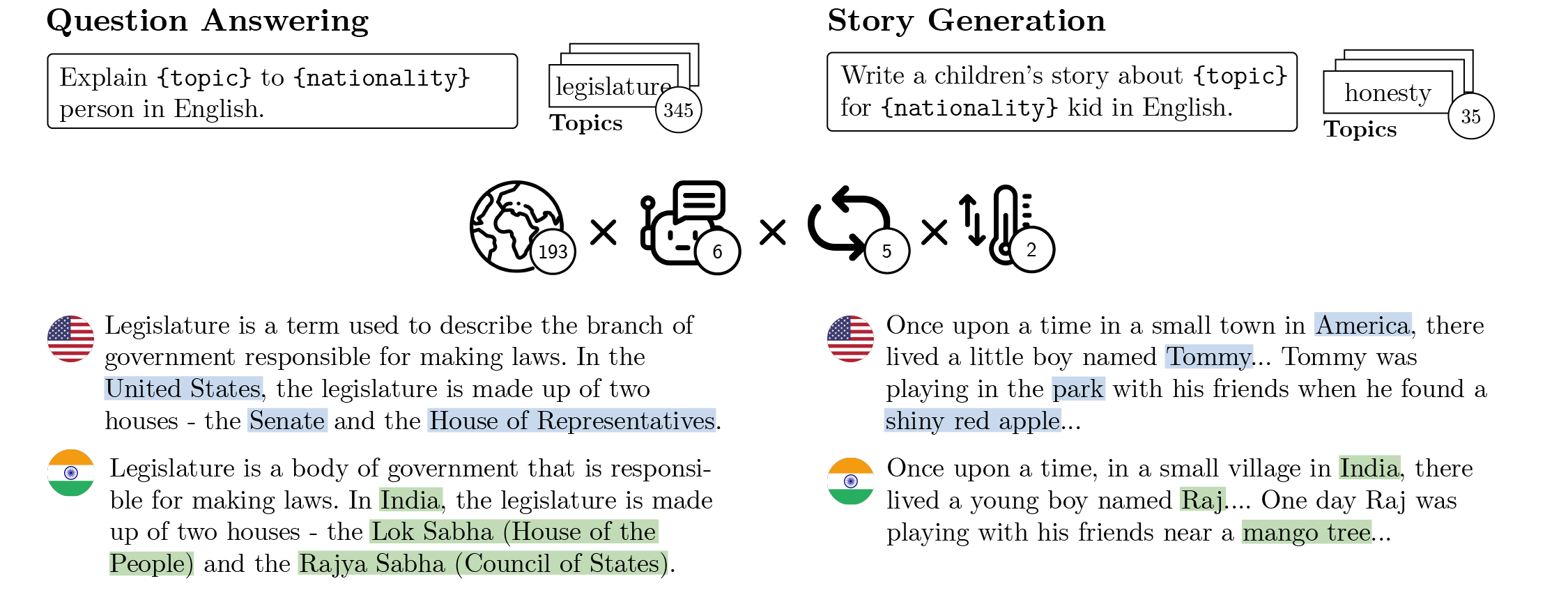}
    \caption{Outputs for Question Answering and Story Generation vary when explicit cue of culture, i.e. nationality, in the prompt is perturbed. We collect outputs for 345 topics for QA, 35 topics for stories, 193 nationalities, 6 LLMs, 5 responses per prompt, and 2 temperatures. We then evaluate these outputs for the extent of lexical variance (\S \ref{sec:5.1-results-lexical-variance}), culturally relevant vocabulary (\S \ref{sec:5.3-results-words}), and correlation between text distribution and the cultural values (\S \ref{sec:5.2-results-kendalls-tau}).}
    \label{fig:eecc}
\end{figure*}

\textit{Cultural competence} is the ability to effectively and appropriately communicate with socioculturally different audiences \cite{deardorff_sage_2009}.
People demonstrate cultural competence by tailoring their utterances to the participants in a conversation \cite{Bell_1984, hawkins_respect_2021, wu_group_2023}. These adaptations range from sociolinguistic variations (e.g., using `soccer' or `football' depending on the context) to appropriately using facts (e.g., in India, the Prime Minister is the head of the government, but in the USA, the President is). 
Hence, for effectively serving diverse users, outputs from large language models (LLMs) need to be culturally relevant \cite{hovy-yang-2021-importance}.

Cultural competence includes a person's \textit{knowledge} of a culture, which then supplements their \textit{skills} of effectively communicating with people from that culture \cite{deardorff_identification_2006,fantini_exploring_2006,alizadeh_cultural_2016}. 
So, the cultural competence of LLMs should also be evaluated along both these aspects.
Contemporary works have largely targeted the \textit{knowledge} component of cultural competence by evaluating LLMs' knowledge of cultural values, norms, and artefacts (\S~\ref{sec:2.2-intrinsic-eval}). Such evaluation is \textit{intrinsic} because it is decoupled from the manifestation of this knowledge in downstream applications \cite{Jones1995}.

In this work, we focus on \textit{extrinsic} evaluation of cultural competence.
Extrinsic evaluation setups should closely mimic user interactions with a system \cite{Jones1995}. 
We select the tasks of story generation and open-ended question answering (QA), both of which have high representation in user interactions with LLMs \cite{zhao2024wildchat}, so . As shown in Figure \ref{fig:eecc}, we obtain model outputs for 35 topics in story generation and 345 topics for QA when an explicit cue of culture, i.e., nationality, is present in the prompt.
Our data consists of outputs from 6 LLMs for 193 nationalities, with 5 outputs per prompt and 2 temperature settings, resulting in over 370K story outputs and 3.6M QA outputs. We analyse the extent and the nature of lexical variations in these outputs.
Further, recent intrinsic evaluations have heavily relied on surveys from cross-cultural psychology, like Hofstede's Cultural Dimensions \cite{Hofstede2010CulturesAO} and World Values Survey \cite{wvs2022} as measures of cultural values (e.g., \citet{arora-etal-2023-probing,durmus2023measuring,alkhamissi2024investigating}, \textit{inter alia}). We evaluate whether the text distributions of outputs correlate with the cultural values of countries, as captured by these surveys.
To summarize, our main research questions are:

\begin{enumerate}[label=\textbf{RQ\arabic*:}, leftmargin=*, parsep=0cm, itemsep=0.25em]
\item Do models vary outputs when explicit cues of culture are present in the input prompt?
\item Do model outputs contain culturally \\relevant vocabulary?
\item Are model outputs for countries with similar cultural values, also similar? 
\end{enumerate}

By measuring the variance in the outputs, we find that models make non-trivial adaptations for different nationalities (\S~\ref{sec:5.1-results-lexical-variance}). Next, inspecting the vocabulary of these outputs, we find that they contain culturally relevant words (\S~\ref{sec:5.3-results-words}). Finally, we find only a weak correlation between the text distributions and cultural values of countries, as measured by cross-cultural psychology surveys frequently used in contemporary work (\S~\ref{sec:5.2-results-kendalls-tau}).

Our findings show that intrinsic and extrinsic measures of cultural competence do not correlate. This necessitates developing holistic evaluations to analyse cultural competence in tasks representative of user interactions with LLMs.

We make our \href{https://github.com/shaily99/eecc}{code}\footnote{\url{https://github.com/shaily99/eecc}} and \href{https://huggingface.co/datasets/shaily99/eecc}{data}\footnote{\url{https://huggingface.co/datasets/shaily99/eecc}} publicly available for replicating results and use in further analysis. 
\section{Related Work}
\label{sec:2-related-work}

\subsection{Cultural Competence}

\textit{Cultural competence}\footnote{interchangeably, the terms intercultural competence and cross-cultural competence are also used.} 
is the ability to effectively communicate with a socioculturally different audience \cite{deardorff_sage_2009}. While multiple definitions exist \cite{alizadeh_cultural_2016}, agreed-upon components include \begin{inlinelist}
    \item the \textit{awareness} about one's positionality and attitude,
    \item the \textit{knowledge} about the language, values, beliefs, practices, symbols, etc. of a culture, and 
    \item the \textit{skill} of appropriately using this \textit{knowledge} when communicating 
\end{inlinelist}\cite{HowardHamilton1998PromotingME,deardorff_identification_2006,fantini_exploring_2006,deardorff_sage_2009}.\footnote{For LLMs, we only rely on analogy to `knowledge' and `skills', and do not invoke analogies to `awareness'.} 

The \textit{knowledge} component requires understanding differences in values, beliefs, and preferences across societies. Surveys in cross-cultural psychology, like Hofstede's Cultural Dimensions (HCD) \cite{Hofstede2001CulturesCC} and World Values Survey (WVS) \cite{wvs2022} attempt to elicit these differences across cultures, proxied by nationalities, using value-based questions.\footnote{For example one of the questions in the Hofstede's survey is ``In choosing an ideal job, how important would it be to you to have sufficient time for your personal or home life?''. }
Survey responses from a large number of individuals are used to quantify the differences in cultural values across countries.
Hofstede's theory, in particular, has been widely adopted in fields requiring cultural competence such as communication, education, business, and healthcare \cite{Ahern2012LostIT,Burai2016SubstructureOA,Chang2023TheCB,Singh2023DigitalTI}. 
\footnote{We note that defining the underpinning concept of culture itself remains elusive. Numerous works have attempted to synthesize the definitions of culture across disciplines, highlighting its complex and multi-faceted nature \cite{culture_kk_book,Baldwin2006AMT}. Broadly, culture is a shared collection of knowledge, values, practices, norms, and beliefs that manifest in expression as behavioural and linguistic patterns \cite{culture_kk_book}.}

\subsection{Cultural Competence in LLMs}
\label{sec:2.2-intrinsic-eval}
There is a growing body of work on ensuring that LLMs align with diverse human values \cite{hershcovich-etal-2022-challenges, wu-etal-2023-cross, kirk2024prism, sorensen2024roadmap} and can serve socioculturally diverse users \cite{hovy-yang-2021-importance,hershcovich-etal-2022-challenges,adilazuarda2024measuring}. 
Specifically, prior works have evaluated LLMs for:

\textit{1. Reflection of diverse cultural values} on cross-cultural psychology surveys (like HCD and WVS) using MCQs, Chain of Thought prompting, or personas \cite{arora-etal-2023-probing,cao-etal-2023-assessing,durmus2023measuring,ramezani2023knowledge,alkhamissi2024investigating,masoud2024cultural}.

\textit{2. Knowledge about varying norms} in social settings like dining, gifting, etc., using yes-no questions \cite{dwivedi-etal-2023-eticor}, natural language inference  \cite{huang-yang-2023-culturally}, red-teaming \cite{chiu2024culturalteaming}, situational questions  \cite{rao2024normad, shi2024culturebank}, and graphs \cite{acharya2020atlas}.

\textit{3. Commonsense and figurative language understanding} using MCQs \cite{palta-rudinger-2023-fork,kabra2023multilingual,kim2024click,koto2024indoculture,wang2024seaeval}, and pragmatic games  \cite{shaikh-etal-2023-modeling}.

\textit{4. Information about cultural artefacts} like food, clothing, etc. \cite{li2024culturegen,seth2024dosa}.

These works reveal gaps in LLMs' knowledge of non-western cultures, complementing known demographic biases in LLMs \cite{mishra2020assessing,zhou2022richer,basu2023inspecting,jha-etal-2023-seegull,schwöbel2023geographical,naous2024having}.

These evaluations focus on the \textit{knowledge} component of cultural competence and are \textit{intrinsic} because they are decoupled from the manifestation of this knowledge in user-facing tasks. Our work is complementary as we evaluate cultural competence in the \textit{extrinsic} setup of text generation.

\section{Extrinsic Evaluation of Cultural Competence}
\label{sec:3-extrinsic-evaluation}

\citet{Jones1995} describe \textit{extrinsic} evaluation criteria as, ``\textit{those relating to its function, i.e its role in relation to its setup's purpose}''. 
So, consider the two broad use cases of LLMs: \begin{inlinelist}
    \item classification tasks, and 
    \item generation tasks.
\end{inlinelist} 
While incorporating cultural knowledge has been shown to benefit classification tasks like hate speech detection and commonsense reasoning \cite{zhou-etal-2023-cultural,li2024culturellm,shi2024culturebank}, to the best of our knowledge there is no prior work focusing on open-ended text generation tasks. 

Specifically, we obtain model outputs when nationalities in prompts are perturbed. We propose quantitative (\S~\ref{sec:3.1-quant-eval}) and qualitative (\S~\ref{sec:3.2-qual-eval}) analyses to evaluate these outputs for cultural competence. 

\subsection{Quantitative Evaluation}
\label{sec:3.1-quant-eval}
We evaluate outputs quantitatively in two ways:

\paragraph{Lexical Variance}
In order to quantify how much the generated language varies when nationalities are perturbed, we measure the variance in word edit distance between outputs.  

\paragraph{Correlation with Cultural Values}
Prior works have relied on cultural values measured by surveys like HCD and WVS for intrinsic evaluation of cultural competence \cite{arora-etal-2023-probing,cao-etal-2023-assessing,durmus2023measuring,ramezani2023knowledge,alkhamissi2024investigating,masoud2024cultural}. So, we evaluate whether the text distributions of outputs correlate with distributions of cultural values. The intuition is to analyse whether countries with similar cultural values have similar text outputs.

We use the \kt rank correlation for this analysis. For each nationality (called the anchor), we rank all other countries by: \begin{inlinelist}
    \item the similarity between their output to the output of the anchor, and
    \item the difference in their cultural values and that of the anchor
\end{inlinelist} to the anchor. We use \kt to calculate the rank correlation of these rankings.

\subsection{Qualitative Evaluation}
\label{sec:3.2-qual-eval}

The qualitative evaluation is intended to assess the characteristics of the outputs when the nationalities are perturbed. 
For this, we inspect the vocabulary of the LLM outputs by surfacing words that occur more frequently in the outputs of a particular country. We used the TF-IDF statistic to obtain words highly relevant to a particular country. The outputs were first tokenized using NLTK \cite{nltk}. Then, we created term frequency vocabulary of all the unigrams occurring in the outputs for each country, considering all outputs of a country as a single `document'. We then calculate the TF-IDF score for all these unigrams and manually inspect the top 15 words for a subset of countries.
\begin{figure*}[ht]
    \centering
    \begin{subfigure}[b]{0.48\textwidth}
        \centering
        \includegraphics[width=\textwidth]{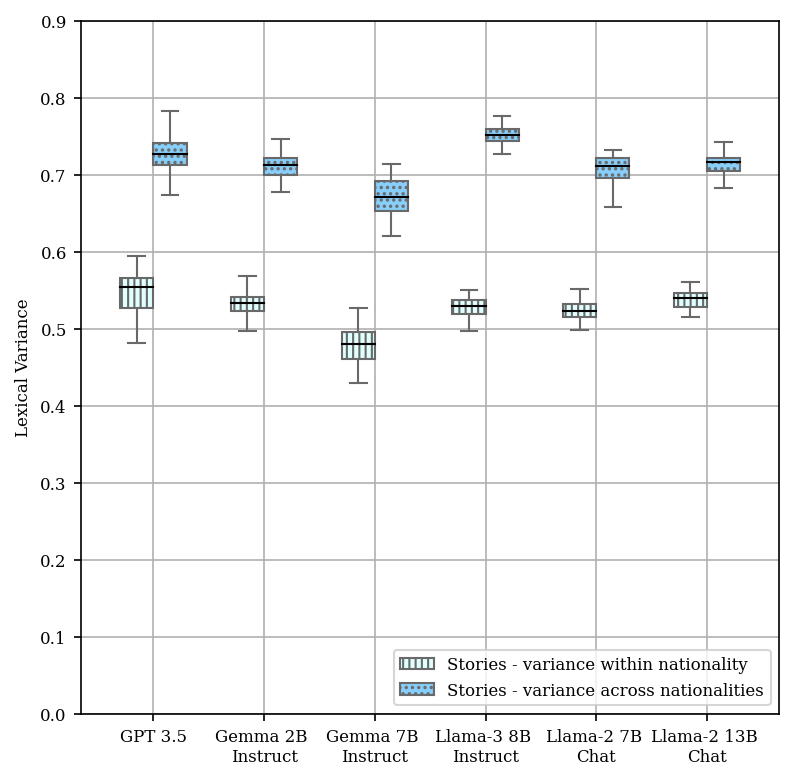}
        \caption{Story Generation}
        \label{fig:lv_stories}
    \end{subfigure}
    \hfill
    \begin{subfigure}[b]{0.48\textwidth}
        \centering
        \includegraphics[width=\textwidth]{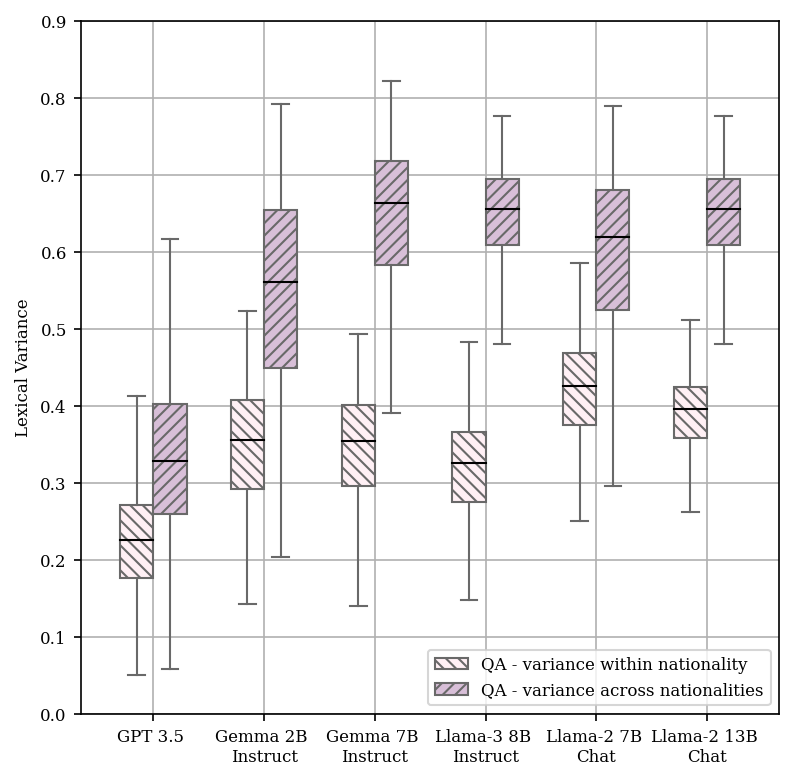}
        \caption{Question Answering}
        \label{fig:lv_qa}
    \end{subfigure}
    \caption{Lexical Variance in outputs. The variance of outputs across nationalities is consistently higher than the variance of outputs within nationalities. Story generation has a higher median variance than QA across models.}
    \label{fig:lexical-variance}
\end{figure*}

\section{Experimental Setup}
\subsection{Tasks and Data}
We select two tasks, story generation and open-ended question answering for our experiments. These were selected as they fulfil two main criteria. First, they have a sizeable representation in user interactions with LLMs \cite{zhao2024wildchat}. Second, they represent diverse types of generation tasks with story generation on the creative end of the spectrum, while question answering being on the factual end of the spectrum.

\paragraph{Open-Ended Question Answering (QA)}
We created a list of 345 topics across 13 categories. We selected the categories (biology, chemistry, economics, environment, humanities, history, law, maths, physics, politics, religion, space, and world affairs) to ensure diversity in topics. Next, we curated topics for each category by referring to \begin{inlinelist}
    \item textbooks used for ACT and SAT Prep in the US, and NCERT textbooks that are used commonly in high schools in India; and 
    \item index terms of ``topics'' of various categories that were available on Wikipedia and Britannica.
\end{inlinelist}\footnote{The datasheet \cite{gebru2021datasheets} is in Appendix \ref{app:datasheet}.}
Examples of topics include: `elections' in `politics', `inertia' in `physics', `photosynthesis' in `biology'. For this task, use a simple prompt template: 

\vspace{5pt}
\noindent `Explain \texttt{\{topic\}} to a/an \texttt{\{nationality\}} person in English.'

\vspace{5pt}
\noindent These results in prompts like `Explain elections to an Indian person in English.'\footnote{We observed that not including the phrase `in English' in the prompt resulted in GPT 3.5's output often being in the dominant language of the particular country, for example for `Mexican' the output is in Spanish. While this is an interesting phenomenon, analyzing this is beyond the scope of this paper.}

\paragraph{Story Generation}

We created a list of 35 topics for children's stories. We used online websites of to curate the list, specifically, freechildrenstories.com, storyberries.com, and parenteducate.com. Examples include topics like moral values (`honesty', `kindness'), characters (`farm animals', `birds'), and places (`school', `jungle'). Similar to QA, we use a simple prompt template: 

\vspace{5pt}
\noindent `Write a children's story about \texttt{\{topic\}} for a/an \texttt{\{nationality\}} kid in English.'

\vspace{5pt}
\noindent This results in prompts like `Write a children's story about honesty for a Japanese kid in English.'

\subsection{Models}
\label{sec:4.2-models}
We evaluate the following LLMs:
\begin{inlinelist}
    \item GPT 3.5 Turbo (\texttt{gpt-3.5-turbo-0125}),\footnote{\url{https://platform.openai.com/docs/models/gpt-3-5-turbo}} queried via API between February 23 and March 28 2024. 
    \item Gemma 2B instruct and 7B instruct \cite{gemmateam2024gemma}
    \item Llama 2 7B chat and 13B chat \cite{touvron2023llama}
    \item Llama 3 8B instruct \cite{llama3modelcard}
\end{inlinelist}.  We sample 5 responses per prompt, with a temperature of 0.3 and 0.7 (for all except GPT 3.5). We generate a maximum of 100 tokens for QA outputs and 1000 tokens for story outputs.

\subsection{Metrics}
\label{sec:4.3:metrics}

\subsubsection{Text Similarity}
\label{sec:4.3.1-text-similarity-metrics}

\paragraph{BLEU} BLEU \cite{papineni-etal-2002-bleu} 
calculates the precision of the n-grams present in the model-generated candidate text as compared to a gold reference text. We re-purpose this to calculate the similarity between two outputs.
Because BLEU is not symmetric, we take the average of the two possible BLEU scores, one with each of the outputs as a candidate and the other as a reference.

\paragraph{Word Edit Distance (WED)} WED is word-level Levenshtein distance \cite{levenshtein}, normalized by the length of the longer text.

\vspace{5pt}
\noindent We picked BLEU and WED to focus on capturing the differences in lexical items between two outputs, e.g., the use of `soccer' or `football'.\footnote{In early experiments we found that  semantic metrics like BERTscore \cite{bertscore} or embedding similarity might not be suitable because: \begin{inlinelist}
    \item a lot of culturally relevant words from the outputs were converted to [UNK] tokens,
    \item we did not see differences in the embedding for outputs that were qualitatively different, especially in QA; perhaps partly because of (a) and because, intuitively the the different words convey the same meaning. 
\end{inlinelist}}

\subsubsection{Difference in Cultural Values}
\label{sec:4.3.2-cultural-similarity-metrics}

Following prior work, we rely on data from cross-cultural psychology surveys to measure the differences in cultural values among countries.

\paragraph{Hofstede's Cultural Dimensions (HCD) } 
Hofstede's cultural theory quantifies the culture of a country along 6 dimensions. Using the VSM2013 version of the data available for 94 countries, we represent each country with 6 dimensions.\footnote{\url{https://geerthofstede.com/research-and-vsm/dimension-data-matrix/}}

\paragraph{World Values Survey (WVS)} We use data from 64 countries and represent each country with 249 dimensions using the 249 questions from WVS.\footnote{There are additional questions that are either non-ordinal or descriptive in nature or are experimental, which we ignore.}\footnote{\url{https://www.worldvaluessurvey.org/WVSDocumentationWV7.jsp}}

\vspace{5pt}
\noindent We calculate the distance in cultural values between two countries as the magnitude of the vector distance between their HCD or WVS representations.

\begin{table*}[h]
\small
\centering
\begin{tabular}{ll}

\toprule
\textbf{Nationality} & \textbf{Top 15 highest TF-IDF scoring words for GPT 3.5's outputs of Story Generation}\\  \midrule
Afghan    & amir,  ali,  afghanistan,  ahmad,  zahra,  amina,  rostam,  babar,  sara,  omar,  \\& cally,  farid,  afghan,  treehouse,  bari    \\\midrule
American  & tommy,  lily,  america,  jack,  jake,  buddy,  mommy,  max,  town,  daddy,  \\& acres,  sarah,  finley,  assignment,  surgery \\\midrule
British   & oliver,  england,  jack,  tommy,  lily,  willowbrook,  thomas,  sherwood,  \\ & littleton,  emily,  british,  jones,  merlin,  london,  teddy  \\\midrule
Canadian  & liam,  canada,  emily,  jack,  alex,  sarah,  canadian,  maple,  tim,  lily, \\& beavers,  smith,  sammy,  moose,  robby   \\\midrule
Chinese   & li,  mei,  china,  ming,  chen,  wu,  wei,  xiao,  wukong,  feather,  ping,  lake,  bao,  snowball,  chinese \\\midrule
German    & hans,  germany,  lena,  anna,  fritz,  max,  gretchen,  bauer,  lorelei, herr,  lila,  liesl,  rübezahl,  emma,  karl   \\\midrule
Indian    & raj,  india,  rani,  arjun,  ravi,  priya,  guru,  peacock,  krishna,  raja,  meena, gupta,  durga,  beggar,  temple  \\\midrule
Nigerian  & kola,  nigeria,  tunde,  bola,  kemi,  ade,  oya,  adaeze,  ayo, zuri,  lagos,  jide,  nigerian,  simba,  heron   \\
\bottomrule
\end{tabular}
\caption{Top 15 highest TF-IDF scoring words for GPT 3.5's outputs of story generation for selected countries}
\label{tab:tfidf_stories}
\end{table*}

\begin{table*}[h]
\small
\centering
\begin{tabular}{ll}
\toprule
\textbf{Nationality} & \textbf{Top 15 highest TF-IDF scoring words for GPT 3.5's outputs for `Politics' in QA}  \\\midrule
Afghan        & afghanistan,  jirga,  ballot,  wolesi,  meshrano,  elders,  afghan,  tribal,  \\ &partners,  box,  strategies,  target,  stake,  exploited,  dynamics   \\\midrule
American      & united,  states,  basis,  four,  american,  expanded,  gun,  fundraising,  accent,  \\&congress,  qualifications,  residency,  requirements,  allowed,  register   \\\midrule
British       & uk,  british,  mps,  commons,  reach,  becomes,  five,  earlier,  lords,  scottish,  \\&brexit,  kingdom,  evolved,  socioeconomic,  previously \\\midrule
Canadian & provincial,	municipal	, federal,	age	, levels,	grassroots, 	shapes, 	\\&riding, 	sector, 	aggression, 	canadian,	guaranteed,	ndp	quebec,	ontario \\\midrule
Chinese & royalty,	enacted,	self-interests,	solving,	achieving, 	something	, channels,	box,	health, \\&	directing	, self-governing	, capable	, prosperous	, citizenship,	accumulation,	accomplish \\\midrule
German        & bundestag,  totalitarian,  he,  argued,  precedence,  opposed,  germany,  upholds, \\& notably,  tourism,  showcase,  transition,  mixed,  emerged,  europe     \\\midrule
Indian        & india,  sabha,  lok,  rajya,  linguistic,  lacking,  flexibility,  chance,  violent,  anarch, \\& hindu,  bharatiya,  janata,  bjp,  indian     \\\midrule
Nigerian & nigeria,	guarantees,	figureheads,	progressives,	apc,	pdp,	purely,	senators,	problem,	finances,	\\&identification,	evenly,	leave,	lawlessness,	governors\\
\bottomrule
\end{tabular}
\caption{Top 15 highest TF-IDF scoring words for GPT 3.5's outputs for `politics' in QA for selected countries}
\label{tab:tfidf_politics}
\end{table*}

\begin{figure*}[h]
    \centering
    \begin{subfigure}[b]{0.48\textwidth}
        \centering
        \includegraphics[width=\textwidth]{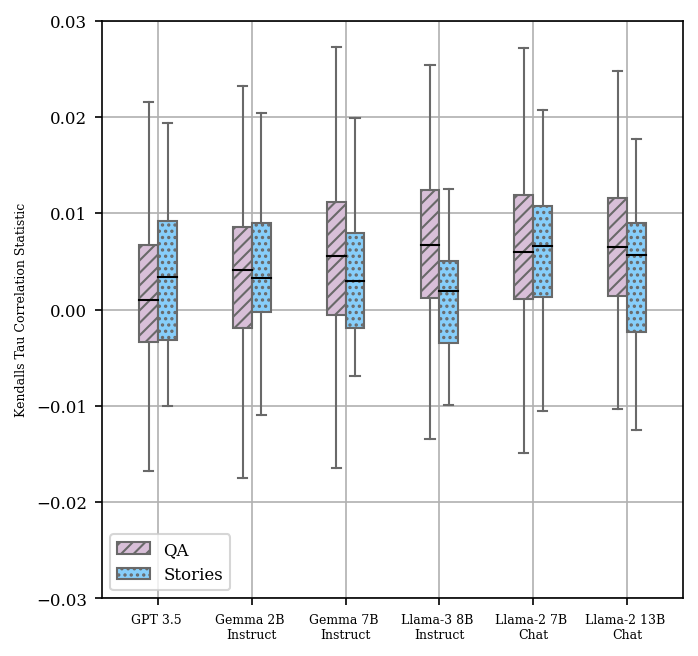}
        \caption{Correlation with Hofstede's Cultural Dimensions (HCD)}
        \label{fig:kt_hcd}
    \end{subfigure}
    \hfill
    \begin{subfigure}[b]{0.48\textwidth}
        \centering
        \includegraphics[width=\textwidth]{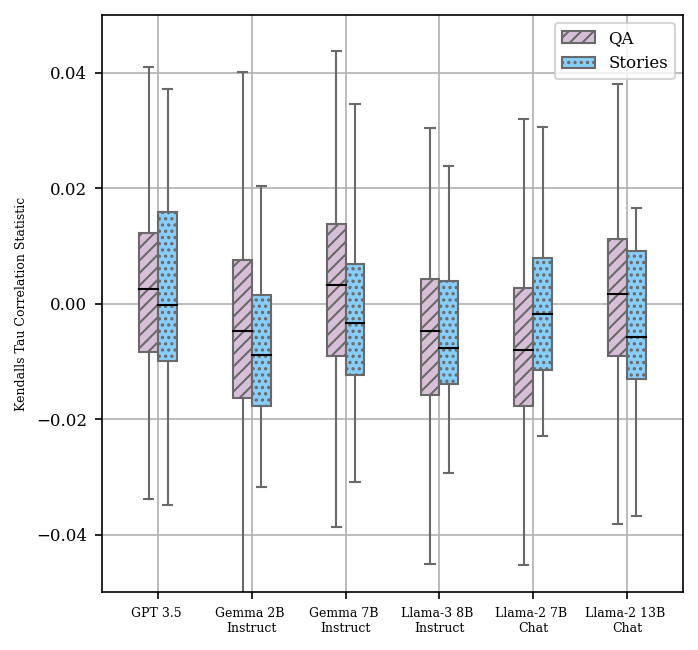}
        \caption{Correlation with World Values Survey (WVS)}
        \label{fig:kt_wvs}
    \end{subfigure}
    \caption{\kt rank correlation between text distribution and cultural closeness of countries. For both plots,  text similarity is measured using \textbf{BLEU}. For HCD correlation statistic values are greater than 0,  implying a small but positive correlation (\ref{fig:kt_hcd}). However, for WVS,  most correlations are less than 0,  indicating small and negative correlation (\ref{fig:kt_wvs}). There are no clear trends among different models or tasks. }
    \label{fig:kt}
\end{figure*}

\begin{figure*}[h]
    \centering
    \begin{subfigure}[b]{0.48\textwidth}
        \centering
        \includegraphics[width=\textwidth]{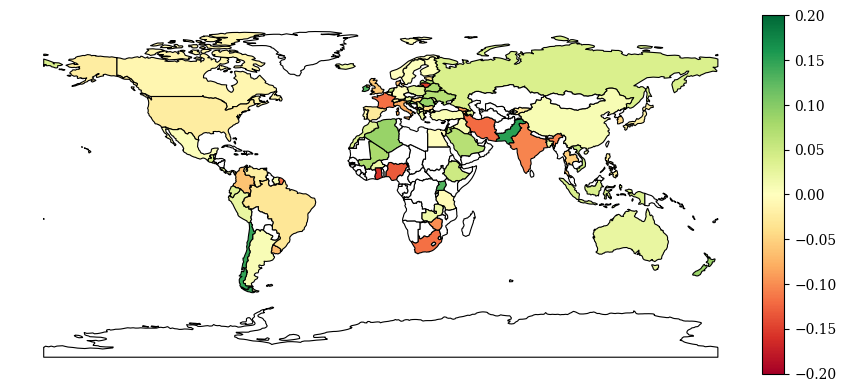}
        \caption{Correlation with Hofstede's Cultural Dimensions (HCD)}
        \label{fig:world_map_hcd}
    \end{subfigure}
    \hfill
    \begin{subfigure}[b]{0.48\textwidth}
        \centering
        \includegraphics[width=\textwidth]{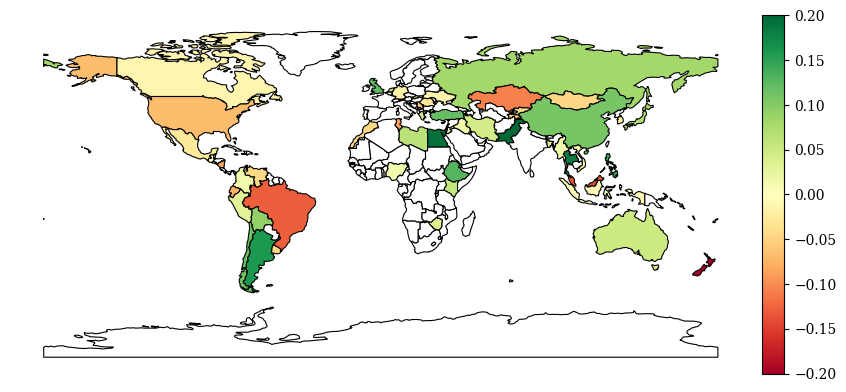}
        \caption{Correlation with World Values Survey (WVS)}
        \label{fig:world_map_wvs}
    \end{subfigure}
    \caption{\kt rank correlation between cultural closeness and text outputs of \textbf{story generation} for GPT 3.5. For both plots, text similarity is measured using \textbf{BLEU}. There is a mix of positive (\textcolor{teal}{green}) and negative (\textcolor{red}{red}) correlation. Russia,  China,  and Australia have positive correlations while India,  USA,  and Canada have negative correlations. European,  South American,  and African countries are split between positive and negative correlations.}
    \label{fig:world_map_story}
\end{figure*}

\section{Results}
We now describe our observations on the outputs obtained with the temperature of 0.3. The observations with a temperature of 0.7 are consistent with these findings and are described in appendix \ref{app:temp7}.

\subsection{Variance Due to Nationality Perturbation}
\label{sec:5.1-results-lexical-variance}

Our first research question was to analyse the extent of variation in outputs when nationalities are perturbed in the prompt. For this, we quantify the lexical variance (\S~\ref{sec:3.1-quant-eval}) in outputs, as measured by word edit distance in Figure \ref{fig:lexical-variance}. We find that model outputs do vary with changing nationalities for both tasks across models. Moreover, these variations are non-trivial and task-dependent, as described below.

\paragraph{Control Experiment: Variance Within Nationalities}
In order to ensure that the variance observed across nationalities is non-trivial, i.e. they do not occur because of the non-deterministic nature of generation in models, we also measure the variance within multiple outputs for a particular nationality. We find that the variance for outputs within nationality is consistently lower than the variance across nationalities. We confirm this with ANOVA having a p-value of <0.05 (Appendix \ref{app:anova}).

\paragraph{Effect of Task on Variance}
We find that the nature of the task affects the extent of variation. The median variance for story generation is higher than the median variance for QA for every model. This might be expected as story generation, had longer outputs and being a creative task allows for more adaptations.
On the other hand,  the difference between the upper and lower quartiles of variance for QA is larger than that for stories. 
This is likely because QA consists of a wider variety of topics ranging from scientific categories, where limited variations might be expected, to politics and history, that allow more variation in answers than others. For example, answers while explaining `elections' (politics) might vary more as they are operationalized differently across countries, but explaining `inertia' (physics) might not vary as much.

\subsection{Culturally Relevant Words in Outputs} \label{sec:5.3-results-words}

Our second research question was to characterize the content of the outputs and understand whether they contain culturally relevant words. For this, we inspected the vocabulary of the outputs. We extracted words highly correlated to a country using TF-IDF (\S~\ref{sec:3.2-qual-eval}). The top 15 words from a subset of countries from outputs of GPT 3.5 for story generation and topics in the politics category from QA are presented in Table \ref{tab:tfidf_stories} and \ref{tab:tfidf_politics},  respectively. 

We see that story generation outputs feature different names across countries.
For example,  `amir' in Afghanistan,  `raj' in India, and `oliver' in Britain. 
Other culturally salient artefacts such as `temple' and `peacock' for Indian,  `bao' in Chinese,   and `london' for UK, etc. also show up in the list.

For the topics in the politics category of the QA task, we see words referring to senate houses and political offices of the countries, for example, `lok sabha' and `rajya sabha' in India, `bundestag' for Germany, and `meshrano jirga' and `wolesi jirga' for Afghanistan. The list also features politically polarised issues such as `gun' in America and `brexit' in UK. Another common feature is the names of political parties, such as `bjp' in India, `apc' and `pdp' in Nigeria, and `ndp' in Canada.\footnote{Proper nouns here appear lower-cased because we lower-cased all model outputs before calculating the statistics.}

Finally, we note that the cultural relevance of all the words on the lists is not obvious (e.g. `notably' in German in Table \ref{tab:tfidf_politics}). Moreover, not all topics in the QA setting surface such interpretable lists of culturally relevant words. Especially the lexicon from scientific topics did not reveal interesting examples when inspecting the top-scoring TF-IDF words. This further compliments our earlier finding of output variations being different across tasks.

\subsection{Correlation in Outputs \& Cultural Values} \label{sec:5.2-results-kendalls-tau}

Our third research question is analysing whether the outputs for countries with similar cultural values are similar. We report the \kt rank correlation (\S~\ref{sec:3.1-quant-eval}), averaged across countries, between BLEU text similarity and distance in cultural values measured by HCD and WVS in Figure \ref{fig:kt}.  

\paragraph{Effect of Measure of Cultural Value Used} When HCD is used as the measure of difference in cultural values (Figure \ref{fig:kt_hcd}),  we find that median correlation across the board\footnote{except QA for Gemma 2B Instruct} is greater than 0. This implies a small but positive correlation between the text distribution and cultural values of countries as measured by HCD. However,  when WVS data is used,  we find a small and negative correlation between text distribution and cultural values as measured by WVS (Figure \ref{fig:kt_wvs}). 
All the rank correlation values 
were statistically significant within a significance interval of 95\% in a two-sided p-test.

\paragraph{Correlation for Different Countries} Next,  we analyse the \kt rank correlation for different countries.
Figure \ref{fig:world_map_story},  shows two example plots for GPT 3.5 for story generation. We find that the correlation for USA,  Canada,  and India (in HCD) is negative,  while that of Russia,  China, Japan, and Australia is positive. South American,  African, Southeast Asian and European countries are split between positive and negative values. 
This is interesting as prior work has found gaps in models' knowledge of non-western cultures (for example \citet{alkhamissi2024investigating, masoud2024cultural}), but we do not see a similar trend. Overall, 
the trend for each country is similar for  HCD and WVS.
\section{Discussion}
\label{sec:6-discussion}
\subsection*{Correlation Between Intrinsic and Extrinsic Metrics of Cultural Competence}

Together the findings for RQ2 (\S~\ref{sec:5.3-results-words}) and RQ3 (\S~\ref{sec:5.2-results-kendalls-tau})  suggest that intrinsic and extrinsic measures of cultural competence are not correlated. On the one hand, model outputs from our extrinsic setup feature culturally relevant words (\S~\ref{sec:5.3-results-words}). On the other hand, the text distributions are only weakly correlated with measures of cultural values widely used in intrinsic evaluations of cultural competence (\S~\ref{sec:5.2-results-kendalls-tau}). Thus, even if an LLM reflects the values of every country perfectly (as prior work measures by Hofstede's Cultural Dimensions or World Values Survey), this ability may not be reflective of cultural competence in downstream tasks.\footnote{Complementary facets of intrinsic and extrinsic evaluation have been observed in multiple settings. For example, there is limited correlation between intrinsic and extrinsic fairness metrics \cite{gonen-goldberg-2019-lipstick-pig,goldfarb-tarrant-etal-2021-intrinsic,cao-etal-2022-intrinsic,delobelle-etal-2022-measuring}, and in intrinsic metrics of language model quality (like perplexity) and downstream task performance \cite{faruqui-etal-2016-problems,dudy-bedrick-2020-words}.}

These findings underscore the importance of extrinsic evaluation of cultural competence.  
We thus believe that future work on advancing cultural competence should focus on tasks reflective of user interactions with language technologies.

\subsection*{Need for Comprehensive Human Evaluation}
Our results show that models adapt to explicit cues of culture with culturally relevant words (\S~\ref{sec:5.3-results-words}). But, it is unclear how this will affect user experience.
In prior work, \citet{Lucy2023OneSizeFitsAllEE} found mixed reactions from users when an email auto-reply system adapted to cues of their identities. Moreover, we do not consider any implicit cues of culture, like dialect or topical differences in queries \cite{kirk2024prism}. Understanding whether model adaptations triggered by implicit and explicit cues of culture are useful or desired by users remains open.

Further, as the qualitative evaluation shows, the output contains names that are typically associated with the ethnic majorities of the country. This is reflective of the biases of the models, which can also lead to potentially offensive, and hurtful generations. While user-facing LLMs might have some, albeit imperfect, safeguards against generating outright toxic content, they might still generate stereotypical text for marginalized groups and cause representational harms \cite{gadiraju2023wouldn}.

Thus, the design of extrinsic evaluation of cultural competence should be task-grounded and user-centred.  
Future work should look into designing human evaluation that considers context (when are adaptations useful?), user agency (do users want adaptations?), and representational harms (who is depicted and how?) in a holistic manner.

\subsection*{Accounting for the Multi-faceted, Intersectional, and Dynamic Nature of Culture}

We find that the correlation between text similarity and cultural values is affected by the measure of the cultural values (\S~\ref{sec:5.2-results-kendalls-tau}). One of the reasons for this might be that measures of cultural values like HCD and WVS are imperfect and incomplete.
This is because there are ample disagreements on the very definition of culture \cite{Baldwin2006AMT}. In fact, Hofstede's Cultural Dimension Theory has been widely criticized for its static nature and over-simplification of culture \cite{Signorini2009DevelopingAF}.
Even so, evaluating cultural competence in LLMs heavily relies on these measures of culture, inheriting these flaws. Future work should consider diverse and complementary measures of culture.

Further, like Hofstede's theory, most evaluations of cultural competence are also done using static benchmarks. However, the world is an evolving place where cultural norms and values are not static. They change and develop through complex interactions among societies.
Future work should focus on incorporating evaluation methods like dynamic benchmarking \cite{kiela2021dynabench} or dealing with disagreements \cite{davni2022dealing}, among others to account for the evolving nature of culture.

Finally, in our work, we use nationality as a proxy for culture. Our choice was motivated by the availability of data on cultural values for countries and by similar operationalization in prior work.  However, culture cannot be anchored by nationalities alone. Moreover, countries are not monoliths and comprise of many and diverse communities. 
Calls for inclusive evaluations of fairness in language technologies \cite{Bhatt2022RecontextualizingFI} have led to important recent work on building fairness resources with participatory design \cite{dev-etal-2023-building,dev2023building}. We believe that methods of evaluation of cultural competence should also similarly embrace participatory and intersectional design.

Overall, the holistic evaluation of cultural competence should account for the multi-faceted, intersectional, and dynamic nature of culture.

\section{Limitations} 
While our work serves as a starting point and a call to focus on the extrinsic evaluation of cultural competence, it is not free of limitations. 

First, we perform limited qualitative evaluation, and we do not perform any comprehensive human evaluation of the outputs. We describe considerations for comprehensive human evaluation in \S~\ref{sec:6-discussion}. 

Secondly, our work is anchored on nationalities and relies on imperfect measures of cultural values. 
However, as we describe in detail in \S~\ref{sec:6-discussion}, evaluation of cultural competence demands participatory and intersectional approaches, in addition to accounting for imperfect and static measures of cultures. 

Further, our evaluation of the outputs does not reflect their pragmatic correctness. In other words, have not evaluated whether a model's adaptations for a particular question (eg. `Explain elections...') correctly reflect how the topic is operationalized in the country. Such evaluation needs either expert knowledge or a comparison with verified sources. 

Moreover, in measuring the characteristics of the text distributions, we focus only on vocabulary. This provides a starting point for cultural competence. However, culturally sensitive text will need to be evaluated for further characteristics also, for example adhering to the tonality, formality, or other stylistic expectations that might vary culturally.

Finally, in our evaluation, we prompt the model with the nationality explicitly and in English. However, there might be other implicit cues of culture that trigger adaptations such as the language and dialect of interaction, and topical differences in queries which we do not account for in this work.

We hope that future work can address these limitations to holistically evaluate LLMs for cultural competence in user-facing tasks.



\section{Conclusion}
In this work, we evaluated cultural competence in two tasks, story generation and open-ended question answering. 
Our data contributions include a hand-curated list of 345 diverse question-answering topics and 35 story-generation topics. We also obtain model outputs for 6 models and 193 nationalities which we will make available for further analysis. Our methodological contributions include conceiving two quantitative and one qualitative analysis for the evaluation of LLM outputs for cultural competence. Using these methods, we find that models do vary their outputs with varying nationalities (\S~\ref{sec:5.1-results-lexical-variance}), outputs contain culturally relevant artefacts (\S~\ref{sec:5.3-results-words}), and model outputs weakly correlate with cultural values (\S~\ref{sec:5.2-results-kendalls-tau}). Our findings underscore the importance of comprehensive extrinsic evaluation of cultural competence.

\section*{Ethical Considerations}
\paragraph{Broader implications and Social Impact} We do not study any sensitive content in this paper, but we note that the outputs of the models could have potentially sensitive and offensive content. Further, the cultural competence of LLMs (or lack thereof) can lead to varying experiences for users from different demographic backgrounds. We discuss the importance of considering user agency and representational harms in this context in \S~\ref{sec:6-discussion}. We note that the examples used in the paper are informed by the knowledge and lived experiences of the authors.

\section*{Acknowledgements}
We thank Saujas Vaduguru, Lucy Li, Arjun Subramonian, Anjali Kantharuban, Jessica Huynh, Alfredo Gomez, To Eun Kim, Athiya Devyani, Simran Khanuja, Akhila Yerukola, Jeremiah Milbauer, Maarten Sap, Yulia Tsvetkov, and Sunipa Dev for their valuable feedback throughout the progress of the work. Special thanks to Daniel Vosler, Dan Griffin, and Graham Neubig for their help and administration of the cluster compute.

\bibliography{custom}

\appendix

\section{Lexical Variance}

\subsection{Calculation Details}

The Variance between two discrete random variables can be defined as:

$$
\operatorname{Var}(X)=\frac{1}{n^2} \sum_{i=1}^n \sum_{j=1}^n \frac{1}{2}\left(x_i-x_j\right)^2
$$

Within this equation, $x_i - x_j$ essentially represents the distance between the two points, which we replace with lexical distance or the Word Edit Distance (WED). Thus, repurposing the above variance equation, lexical variance in outputs across nationalities for a concept can be calculated as:

$$
\begin{aligned}
& \frac{1}{|\mathcal{N}|^2} \sum_{n \in \mathcal{N}} \sum_{n^{\prime} \in \mathcal{N}} \frac{1}{2} (\operatorname{WED}(O_{n}, O_{n'}))^2 \\
\end{aligned}
$$

Where: 
$\mathcal{N}$ = Set of all Nationalities, $n$ = nationality, $O_n$ = output for nationality $n$

\subsection{ANOVA results on within and across nationality lexical variance}
\label{app:anova}
\paragraph{}H0 = $\mu_{within} = \mu_{across}$
\paragraph{}H1 = they are different

The p-values are in table \ref{tab:anova}

\begin{table*}[h]
\begin{tabular}{lllll}
\\\hline
\textbf{task} & \textbf{model} & \textbf{F-statistic} & \textbf{p-value} & \textbf{Reject H0}  \\\hline
stories & llama2\_7B\_chat & 2255.3456 & 7.043450519806435e-54 & \textcolor{teal}{Yes} \\
stories & llama2\_13B\_chat & 2821.1494 & 4.248487694091554e-57 & \textcolor{teal}{Yes} \\
stories & llama3\_8B\_instruct & 3610.5356 & 1.1491538258492085e-60 & \textcolor{teal}{Yes} \\
stories & gemma2B\_it & 874.1556 & 1.5311181671386628e-40 & \textcolor{teal}{Yes} \\
stories & gemma7B\_it & 1721.6872 & 5.048199426344931e-50 & \textcolor{teal}{Yes} \\
stories & gpt\_3-58 & 1055.6979 & 3.80594818701481e-43 & \textcolor{teal}{Yes}  \\\hline
QA & llama2\_7B\_chat & 911.4229 & 3.7297913016341677e-128 & \textcolor{teal}{Yes} \\
QA & llama2\_13B\_chat & 1444.7691 & 3.4753916948315105e-171 & \textcolor{teal}{Yes} \\
QA & llama3\_8B\_instruct & 2585.3423 & 3.2168642758230666e-235 & \textcolor{teal}{Yes} \\
QA & gemma2B\_it & 550.97 & 5.966032578765733e-90 & \textcolor{teal}{Yes} \\
QA & gemma7B\_it & 1335.7818 & 2.4154064759769195e-163 & \textcolor{teal}{Yes} \\
QA & gpt\_3-5 & 233.4199 & 1.3687492031148516e-45 & \textcolor{teal}{Yes}  \\\hline
\end{tabular}
\caption{One Way ANOVA for within and across nationalities. All p-values suggest that H0 (same means) can be rejected.}
\label{tab:anova}
\end{table*}

\begin{figure*}[ht]
    \centering
    \begin{subfigure}[b]{0.48\textwidth}
        \centering
        \includegraphics[width=\textwidth]{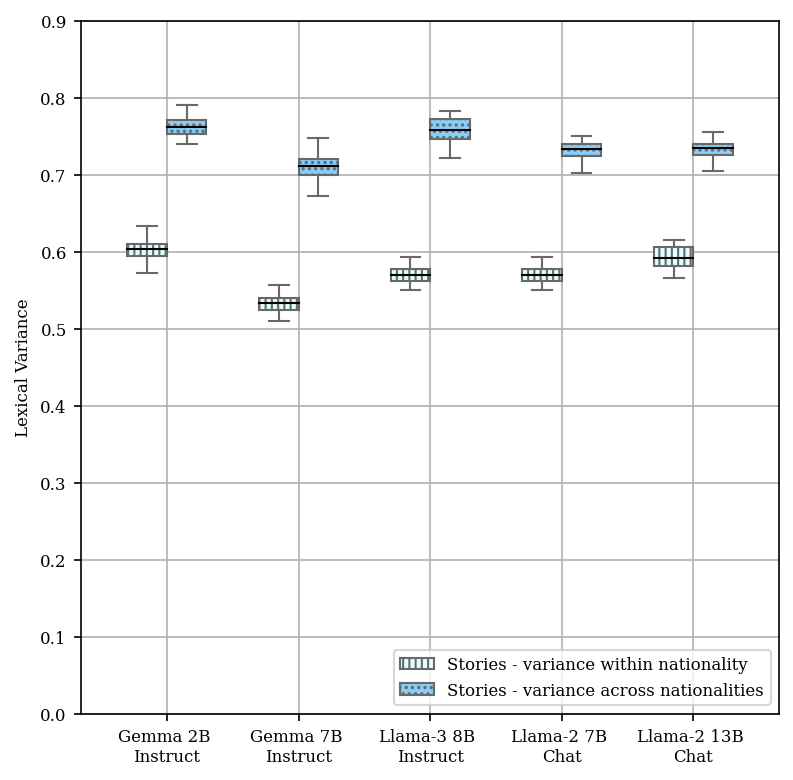}
        \caption{Story Generation}
        \label{fig:lv_stories}
    \end{subfigure}
    \hfill
    \begin{subfigure}[b]{0.48\textwidth}
        \centering
        \includegraphics[width=\textwidth]{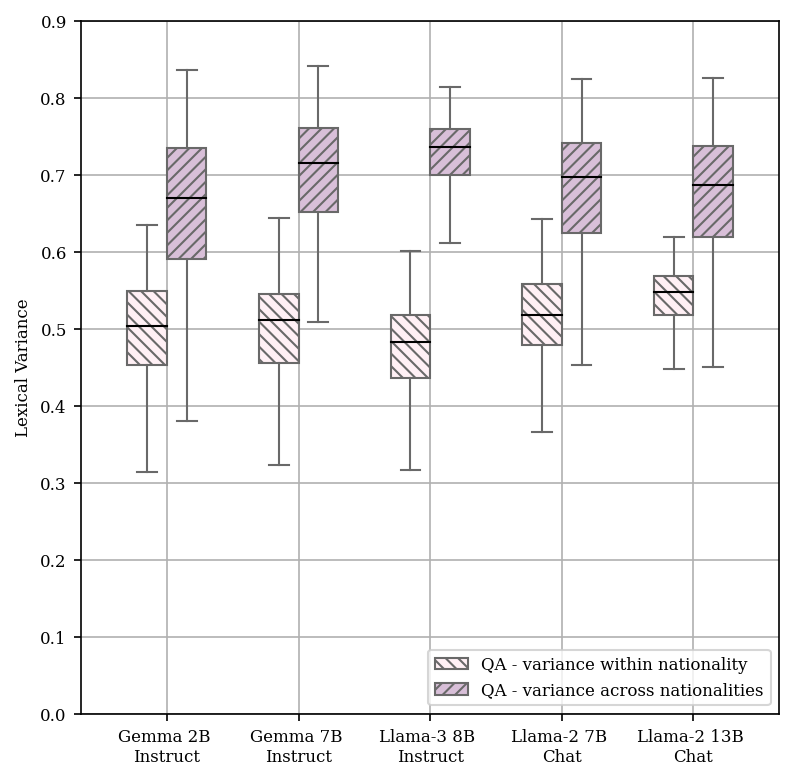}
        \caption{Question Answering}
        \label{fig:lv_qa}
    \end{subfigure}
    \caption{Lexical Variance in outputs with temperature = 0.7. The variance of outputs across nationalities is consistently higher than the variance of outputs within nationalities, as also observed with a temperature of 0.3 in \S \ref{sec:5.1-results-lexical-variance}. Story generation has a higher median variance than QA across models. Note that the absolute values of variances across the board are higher than those obtained for the temperature = 0.3, which is consistent with the expectation of variation in generation increasing with increasing temperature.}
    \label{fig:lexical-variance-7}
\end{figure*}

\section{Kendalls Tau Rank Correlation}

\subsection{Choice of Kendalls Tau variant} We use the c variant in particular because before the ranking both the rank lists have been generated by metrics that have different scales.

\subsection{An example calculation}
This is a brief example of how \kt was calculated. Suppose there are 4 nationalities: A, B, C, D. We first take one nationality as an anchor, let's say A, and create two rank lists. The first rank list is of similarity of text outputs to A, let's say this is [B, D, C] and the second is using the distance between cultural values representation let's say this is [D, C, B]. Here, we reverse the raw rank list we get from distance in the vector representation of cultural values because this was obtained using distance values while the other one was obtained using similarity values,  For A, the rank correlation between these two rank lists (i.e. [B, D, C] and [D, C, B]) is calculated using \kt. We use sklearn to calculate \kt with default parameters. Finally, for a particular concept, we take an average of \kt across all nationalities.

\begin{figure*}[h]
    \centering
    \begin{subfigure}[b]{0.48\textwidth}
        \centering
        \includegraphics[width=\textwidth]{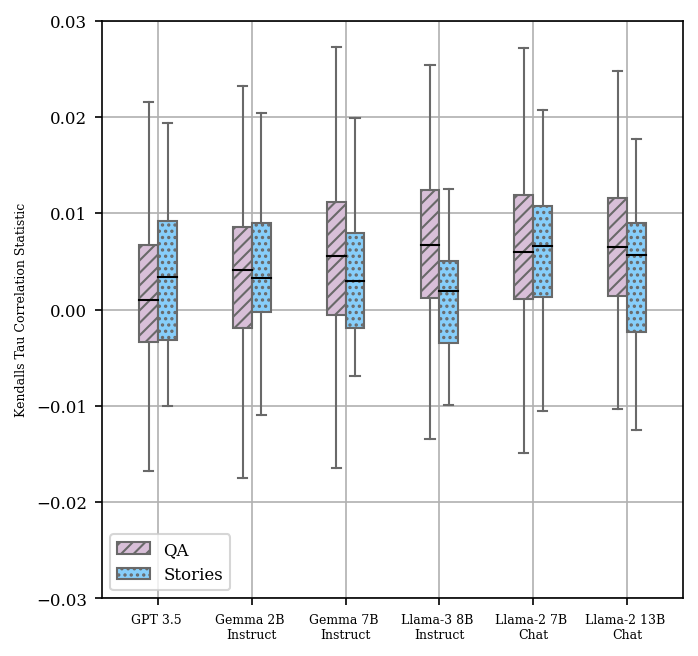}
        \caption{Correlation with Hofstede's Cultural Dimensions (HCD)}
        \label{fig:kt_hcd}
    \end{subfigure}
    \hfill
    \begin{subfigure}[b]{0.48\textwidth}
        \centering
        \includegraphics[width=\textwidth]{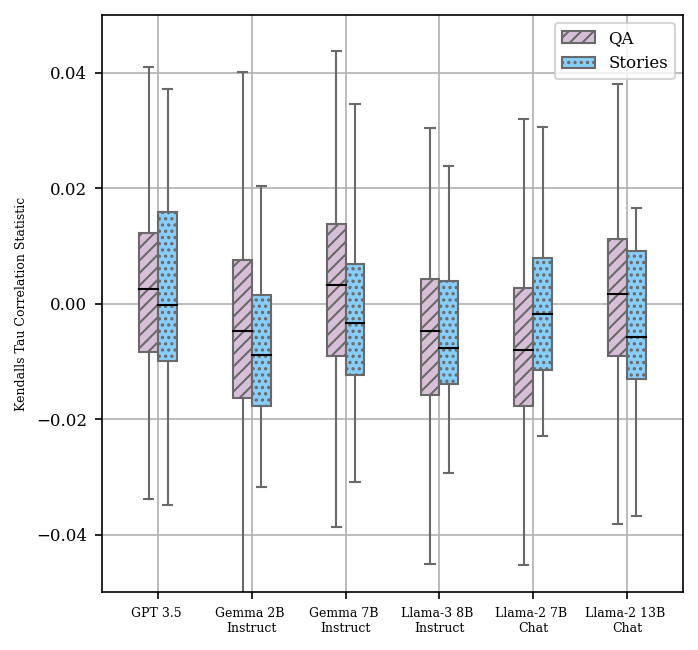}
        \caption{Correlation with World Values Survey (WVS)}
        \label{fig:kt_wvs}
    \end{subfigure}
    \caption{\kt rank correlation between text distribution and cultural closeness of countries. For both plots,  text similarity is measured using \textbf{BLEU}. For HCD correlation statistic values are greater than 0,  implying a small but positive correlation (\ref{fig:kt_hcd}). However, for WVS,  most correlations are less than 0,  indicating small and negative correlation (\ref{fig:kt_wvs}). There are no clear trends among different models or tasks. We note that these findings are consistent with the findings of \S \ref{sec:5.2-results-kendalls-tau}.}
    \label{fig:kt}
\end{figure*}

\section{Results with Temperature = 0.7}
\label{app:temp7}
In this section, we describe the results obtained with outputs when a temperature of 0.7 is used during generation. We note that the findings are consistent with the results described in \S 5, albeit with different absolute values.

Figure \ref{fig:lexical-variance-7} shows the lexical variance for outputs generated with temperature = 0.7. Consistent with findings from \S \ref{sec:5.1-results-lexical-variance}, the median variance across nationalities is higher than the median variance within nationalities, variance in story outputs is consistently more than QA outputs, and the difference between quartiles is higher for QA than for stories. All of this is consistent with findings from \S \ref{sec:5.1-results-lexical-variance}. Note that all the variance values across the board are higher than the variance values obtained with a temperature of 0.3. This is expected as randomness in generations of LLMs is known to increase with increasing temperature.

Next, for correlation between the similarity of model outputs and cultural values among nationality, we also observe consistent findings at a higher temperature. Specifically, similar to findings in \S \ref{sec:5.2-results-kendalls-tau}, all \kt correlation values are close to 0, generally on the +ve side when HCD is used and -ve when WVS is used, and there is no significant difference based on the task.

\section{Datasheet}
\label{app:datasheet}

\noindent

This document is based on \textit{Datasheets for Datasets} by \citet{gebru2021datasheets}. The latex template is based on this \href{https://github.com/AudreyBeard/Datasheets-for-Datasets-Template/tree/master}{github repo}

\subsection{Motivation}

\textit{\textbf{
For what purpose was the dataset created?
}
Was there a specific task in mind? Was there
a specific gap that needed to be filled? Please provide a description.}

This dataset has two parts. First is a list of topics to prompt models with for two tasks, question answering and story generation to analyse differences in model outputs across nationalities. Second are the model responses for these prompts.

\textit{\textbf{
Who created this dataset (e.g., which team, research group) and on behalf
of which entity (e.g., company, institution, organization)?
}
} 

Anonymized for peer review

\textit{\textbf{
What support was needed to make this dataset?
}
(e.g. funded the creation of the dataset? If there is an associated
grant, provide the name of the grantor and the grant name and number, or if
it was supported by a company or government agency, give those details.)
} 

Anonymized for peer review  
    
\subsection{Composition}

\textit{\textbf{
What do the instances that comprise the dataset represent (e.g., documents,
photos, people, countries)?
}
Are there multiple types of instances (e.g., movies, users, and ratings;
people and interactions between them; nodes and edges)? Please provide a
description.
} 

The data consists of a list of topics. The model outputs contain text generated by LLMs.

\textit{\textbf{
How many instances are there in total (of each type, if appropriate)?
}} 

35 topics for story generation and 345 topics for QA. We combine this with 193 nationalities to result in 66,585 QA prompts and 6755 Story prompts. We sample 5 responses for every prompt from 6 LLMs and two temperature settings (except GPT temperature 0.7). This leads to 3,662,175 model outputs for QA and 371,525 model outputs for stories.

\textit{\textbf{
Does the dataset contain all possible instances or is it a sample (not
necessarily random) of instances from a larger set?
}
If the dataset is a sample, then what is the larger set? Is the sample
representative of the larger set (e.g., geographic coverage)? If so, please
describe how this representativeness was validated/verified. If it is not
representative of the larger set, please describe why not (e.g., to cover a
more diverse range of instances, because instances were withheld or
unavailable).
}

This is a hand-curated list of data. It is not exhaustively representative of all possible story generation topics or QA topics. For story generation in particular, we only focus on children's stories. For QA, we attempt to include diverse topics and categories. But we note that these are open-ended tasks and thus the range of topics is very wide to measure exhaustiveness.

\textit{\textbf{
What data does each instance consist of?
}
“Raw” data (e.g., unprocessed text or images) or features? In either case,
please provide a description.
} 

Each instance in the topic list is simply a phrase (unigram or bigram) that is used to create a prompt for Question answering or story generation. Each instance of model output is a paragraph with a maximum of 100 tokens in case of QA and 1000 tokens in case of story generation.

\textit{\textbf{
Is there a label or target associated with each instance?
}
If so, please provide a description.
} 

There are no labels

\textit{\textbf{
Is any information missing from individual instances?
}
If so, please provide a description, explaining why this information is
missing (e.g., because it was unavailable). This does not include
intentionally removed information but might include, e.g., redacted text.
} 

No

\textit{\textbf{
Are relationships between individual instances made explicit (e.g., users’
movie ratings, social network links)?
}
If so, please describe how these relationships are made explicit.
} 

No

\textit{\textbf{
Are there recommended data splits (e.g., training, development/validation,
testing)?
}
If so, please provide a description of these splits, explaining the
the rationale behind them.
} 

All of the data is intended for evaluation, we do not anticipate needing any training or validation splits.

\textit{\textbf{
Are there any errors, sources of noise, or redundancies in the dataset?
}
If so, please provide a description.
} 

No

\textit{\textbf{
Is the dataset self-contained, or does it link to or otherwise rely on
external resources (e.g., websites, tweets, other datasets)?
}
If it links to or relies on external resources, a) are there guarantees
that they will exist, and remain constant, over time; b) are there official
archival versions of the complete dataset (i.e., including the external
resources as they existed at the time the dataset was created); c) are
there any restrictions (e.g., licenses, fees) associated with any of the
external resources that might apply to a future user? Please provide
descriptions of all external resources and any restrictions associated with
them, as well as links or other access points, as appropriate.
} 

It is self-contained.

\textit{\textbf{
Does the dataset contain data that might be considered confidential (e.g.,
data that is protected by legal privilege or by doctor-patient
confidentiality, data that includes the content of individuals’ non-public
communications)?
}
If so, please provide a description.
} 

No

\textit{\textbf{
Does the dataset contain data that, if viewed directly, might be offensive,
insulting, threatening, or might otherwise cause anxiety?
}
If so, please describe why.
} 

No

\textit{\textbf{
Does the dataset relate to people?
}
If not, you may skip the remaining questions in this section.
} 

No

\textit{\textbf{
Does the dataset identify any subpopulations (e.g., by age, gender)?
}
If so, please describe how these subpopulations are identified and
provide a description of their respective distributions within the dataset.
} 

For collecting model outputs, the prompt that we use explicitly mentions a nationality. This is because we want to study the perturbation of the model outputs when nationalities are perturbed in the prompts. Because of this model outputs in the data are likely to contain text that refers to respective nationalities.

\textit{\textbf{
Is it possible to identify individuals (i.e., one or more natural persons),
either directly or indirectly (i.e., in combination with other data) from
the dataset?
}
If so, please describe how.
} 

No

\textit{\textbf{
Does the dataset contain data that might be considered sensitive in any way
(e.g., data that reveals racial or ethnic origins, sexual orientations,
religious beliefs, political opinions or union memberships, or locations;
financial or health data; biometric or genetic data; forms of government
identification, such as social security numbers; criminal history)?
}
If so, please provide a description.
} 

No

\textit{\textbf{
Any other comments?
}} 

No

\subsection{Collection}

\textit{\textbf{
How was the data associated with each instance acquired?
}
Was the data directly observable (e.g., raw text, movie ratings),
reported by subjects (e.g., survey responses), or indirectly
inferred/derived from other data (e.g., part-of-speech tags, model-based
guesses for age or language)? If data was reported by subjects or
indirectly inferred/derived from other data, was the data
validated/verified? If so, please describe how.
} 

The topics were obtained by hand-curation. The authors first created a broad list of 13 categories that were of interest in the evaluation: biology, chemistry, environment, economics, history, humanities, law, maths, physics, politics, space, religion, and world affairs. These categories were selected as intuitive categories of questions in which differences in model outputs might be observed. The authors then referred to textbooks and encyclopedia indexes to sample topics within these categories leading to a total of 345 topics. For stories, the authors first similarly selected three broad categories on which children's stories can be written: moral values, stories with specific characters, and stories with specific settings. They then used online websites and children's story books to come up with topics in these areas creating a list of 35 topics. This is the topic list. Next, these were then used in a simple template `Explain \{topic\} to a / an \{nationality\} person.' for QA and `Write a story about \{topic\} for a / an \{nationality\} kid.' in the story. The resulting prompts were input into 6 LLMs listed in \ref{sec:4.2-models} to obtain model outputs. 5 responses were generated for every output.

\textit{\textbf{
Over what timeframe was the data collected?
}
Does this timeframe match the creation timeframe of the data associated
with the instances (e.g., recent crawl of old news articles)? If not,
please describe the timeframe in which the data associated with the
instances was created. Finally, list when the dataset was first published.
} 

The topic list was curated between November 2023 and January 2024. Model outputs were collected between February and April 2024.

\textit{\textbf{
What mechanisms or procedures were used to collect the data (e.g., hardware
apparatus or sensor, manual human curation, software program, software
API)?
}
How were these mechanisms or procedures validated?
} 

The entire data of the topic list is human-curated. The model outputs are LLMs generated. Some characteristics of the model outputs are evaluated in the paper.

\textit{\textbf{
What was the resource cost of collecting the data?
}
(e.g. what were the required computational resources, and the associated
financial costs, and energy consumption - estimate the carbon footprint.
See Strubell \textit{et al.}\cite{strubell2019energy} for approaches in this area.)
} 

The cost of hand-curating topic lists was about 10 researcher hours. For getting model outputs, A6000 GPUs were used for hosting the LLM to run inference for obtaining model outputs. The total inference cost was about 45 GPU hours. Model outputs from GPT 3.5 cost about 125 USD.

\textit{\textbf{
If the dataset is a sample from a larger set, what was the sampling
strategy (e.g., deterministic, probabilistic with specific sampling
probabilities)?
}
} 

We did not sample.

\textit{\textbf{
Who was involved in the data collection process (e.g., students,
crowd workers, contractors) and how were they compensated (e.g., how much
were crowdworkers paid)?
}
} 

The data was hand-curated by the author and the author queried LLMs for model outputs. No additional personnel was involved.

\textit{\textbf{
Were any ethical review processes conducted (e.g., by an institutional
review board)?
}
If so, please provide a description of these review processes, including
the outcomes, as well as a link or other access point to any supporting
documentation.
} 

No human subjects or crowd workers were involved hence we did not conduct any IRB.

\textit{\textbf{
Does the dataset relate to people?
}
If not, you may skip the remainder of the questions in this section.
} 

No

\textit{\textbf{
Did you collect the data from the individuals in question directly, or
obtain it via third parties or other sources (e.g., websites)?
}
} 

NA

\textit{\textbf{
Were the individuals in question notified about the data collection?
}
If so, please describe (or show with screenshots or other information) how
notice was provided, and provide a link or other access point to, or
otherwise reproduce, the exact language of the notification itself.
} 

NA

\textit{\textbf{
Did the individuals in question consent to the collection and use of their
data?
}
If so, please describe (or show with screenshots or other information) how
consent was requested and provided, and provide a link or other access
point to, or otherwise reproduce, the exact language to which the
individuals consented.
} 

NA

\textit{\textbf{
If consent was obtained, were the consenting individuals provided with a
mechanism to revoke their consent in the future or for certain uses?
}
 If so, please provide a description, as well as a link or other access
 point to the mechanism (if appropriate)
} 

NA

\textit{\textbf{
Has an analysis of the potential impact of the dataset and its use on data
subjects (e.g., a data protection impact analysis)been conducted?
}
If so, please provide a description of this analysis, including the
outcomes, as well as a link or other access point to any supporting
documentation.
} 

NA

\textit{\textbf{
Any other comments?
}} 

NA

\subsection{Preprocessing / Labelling / Cleaning
}

\textit{\textbf{
Was any preprocessing/cleaning/labeling of the data
done(e.g.,discretization or bucketing, tokenization, part-of-speech
tagging, SIFT feature extraction, removal of instances, processing of
missing values)?
}
If so, please provide a description. If not, you may skip the remainder of
the questions in this section.
} 

No

\textit{\textbf{
Was the “raw” data saved in addition to the preprocessed/cleaned/labeled
data (e.g., to support unanticipated future uses)?
}
If so, please provide a link or other access point to the “raw” data.
} 

No cleaning was performed.

\textit{\textbf{
Is the software used to preprocess/clean/label the instances available?
}
If so, please provide a link or other access point.
} 

NA

\textit{\textbf{
Any other comments?
}} 

NA

\subsection{Uses}

\textit{\textbf{
Has the dataset been used for any tasks already?
}
If so, please provide a description.
} 

Yes, the data was used to evaluate the variations in model outputs for varying nationalities in the input prompts for two tasks in order to evaluate cultural competence.

\textit{\textbf{
Is there a repository that links to any or all papers or systems that use the dataset?
}
If so, please provide a link or other access point.
} 

Yes. The data, paper, and code will be open-sourced after peer review.

\textit{\textbf{
What (other) tasks could the dataset be used for?
}
} 

The list of topics could be used for a different task evaluation. The model outputs could be further used to characterize model behaviour in these settings, such as qualitative analysis of outputs, analysis for prescence of biases and so on.

\textit{\textbf{
Is there anything about the composition of the dataset or the way it was
collected and preprocessed/cleaned/labeled that might impact future uses?
}
For example, is there anything that a future user might need to know to
avoid uses that could result in unfair treatment of individuals or groups
(e.g., stereotyping, quality of service issues) or other undesirable harms
(e.g., financial harms, legal risks) If so, please provide a description.
Is there anything a future user could do to mitigate these undesirable
harms?
} 

No. We do note though that the model outputs are generated content from LLMs and might contain toxic, offensive, and stereotypical texts against marginalized communities. We advise discretion on the part of users who choose to further utilise this data for analysis.

\textit{\textbf{
Are there tasks for which the dataset should not be used?
}
If so, please provide a description.
} 

The topic lists should not be treated as an exhaustive list of topics to evaluate cultural competence. The model outputs should not be used as gold standard answers for particular questions or story-generation tasks.

\textit{\textbf{
Any other comments?
}} 

No

\subsection{Distribution}

\textit{\textbf{
Will the dataset be distributed to third parties outside of the entity
(e.g., company, institution, organization) on behalf of the dataset
was created?
}
If so, please provide a description.
} 

We make our \href{https://huggingface.co/datasets/shaily99/eecc}{data} and \href{https://github.com/shaily99/eecc/tree/main}{code} publicly available for replicating results and use in further analysis. 

\textit{\textbf{
How will the dataset will be distributed (e.g., tarball on website, API,
GitHub)?
}
Does the dataset have a digital object identifier (DOI)?
} 
We make our \href{https://huggingface.co/datasets/shaily99/eecc}{data} and \href{https://github.com/shaily99/eecc/tree/main}{code} publicly available for replicating results and use in further analysis. 

\textit{\textbf{
When will the dataset be distributed?
}
} 

We make our \href{https://huggingface.co/datasets/shaily99/eecc}{data} and \href{https://github.com/shaily99/eecc/tree/main}{code} publicly available for replicating results and use in further analysis. 

\textit{\textbf{
Will the dataset be distributed under a copyright or other intellectual
property (IP) license, and/or under applicable terms of use (ToU)?
}
If so, please describe this license and/or ToU, and provide a link or other
access point to, or otherwise reproduce, any relevant licensing terms or
ToU, as well as any fees associated with these restrictions.
} 

We make our \href{https://huggingface.co/datasets/shaily99/eecc}{data} and \href{https://github.com/shaily99/eecc/tree/main}{code} publicly available for replicating results and use in further analysis. 

\textit{\textbf{
Have any third parties imposed IP-based or other restrictions on the data
associated with the instances?
}
If so, please describe these restrictions, and provide a link or other
access point to, or otherwise reproduce, any relevant licensing terms, as
well as any fees associated with these restrictions.
} 

No

\textit{\textbf{
Do any export controls or other regulatory restrictions apply to the
dataset or to individual instances?
}
If so, please describe these restrictions, and provide a link or other
access point to, or otherwise reproduce, any supporting documentation.
} 

No

\textit{\textbf{
Any other comments?
}} 

YOUR ANSWER HERE

\subsection{Maintenance}

\textit{\textbf{
Who is supporting/hosting/maintaining the dataset?
}
} 

Anonymized for peer review.

\textit{\textbf{
How can the owner/curator/manager of the dataset be contacted (e.g., email
address)?
}
} 

Anonymized for peer review.

\textit{\textbf{
Is there an erratum?
}
If so, please provide a link or other access point.
} 

No. The authors can be contacted via email.

\textit{\textbf{
Will the dataset be updated (e.g., to correct labelling errors, add new
instances, delete instances)?
}
If so, please describe how often, by whom, and how updates will be
communicated to users (e.g., mailing list, GitHub)?
} 

This data is unlikely to be updated.

\textit{\textbf{
If the dataset relates to people, are there applicable limits on the
retention of the data associated with the instances (e.g., were individuals
in question told that their data would be retained for a fixed period of
time and then deleted)?
}
If so, please describe these limits and explain how they will be enforced.
} 

NA

\textit{\textbf{
Will older versions of the dataset continue to be
supported/hosted/maintained?
}
If so, please describe how. If not, please describe how its obsolescence
will be communicated to users.
} 

We do not intend to have multiple versions.

\textit{\textbf{
If others want to extend/augment/build on/contribute to the dataset, is
there a mechanism for them to do so?
}
If so, please provide a description. Will these contributions be
validated/verified? If so, please describe how. If not, why not? Is there a
process for communicating/distributing these contributions to other users?
If so, please provide a description.
} 

TBD

\textit{\textbf{
Any other comments?
}} 

NA

\end{document}